\documentclass{article}

\usepackage[preprint]{neurips_2019}

\usepackage[utf8]{inputenc} 
\usepackage[T1]{fontenc}    
\usepackage{hyperref}       
\usepackage{url}            
\usepackage{booktabs}       
\usepackage{amsfonts}       
\usepackage{nicefrac}       
\usepackage{microtype}      
\usepackage{graphicx}

\title{Creating Spoken Dialog Systems in Ultra-Low Resourced Settings}

\author{%
  Moayad Elamin\\
  Carnegie Mellon University\\
  \texttt{melamin@andrew.cmu.edu}\\
  \And
  Muhammad Omer\\
  Carnegie Mellon University\\
  \texttt{maomer@andrew.cmu.edu}\\
  \And
  Yonas Chanie\\
  Carnegie Mellon University\\
  \texttt{ychanie@andrew.cmu.edu}\\
  \AND
  Henslaac Ndlovu\\
  Carnegie Mellon University\\
  \texttt{hndlovu@andrew.cmu.edu}\\
}

\begin{document}

\maketitle
\begin{abstract}
Automatic Speech Recognition (ASR) systems are a crucial technology that is used today to design a wide variety of applications, most notably, smart assistants, such as Alexa. ASR systems are essentially dialogue systems that employ Spoken Language Understanding (SLU) to extract meaningful information from speech. The main challenge with designing such systems is that they require a huge amount of labeled clean data to perform competitively, such data is extremely hard to collect and annotate to respective SLU tasks, furthermore, when designing such systems for low resource languages, where data is extremely limited, the severity of the problem intensifies. In this paper, we focus on a fairly popular SLU task, that is, Intent Classification while working with a low resource language, namely, Flemish. Intent Classification is a task concerned with understanding the intents of the user interacting with the system. We build on existing light models for intent classification in Flemish, and our main contribution is applying different augmentation techniques on two levels \_ the voice level, and the phonetic transcripts level \_ to the existing models to counter the problem of scarce labeled data in low-resource languages. We find that our data augmentation techniques, on both levels, have improved the model performance on a number of tasks. 

\end{abstract}

\section{Introduction}

\begin{figure}[!ht]
    \centering
    \includegraphics[scale=0.4]{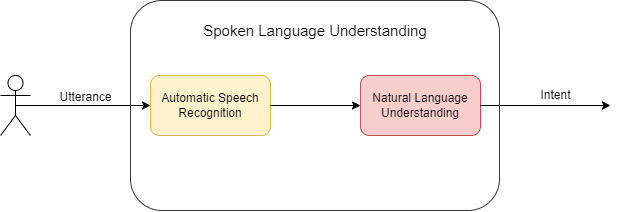}
    \caption{SLU for Intent Recognition}
    \label{fig:model}
\end{figure}

Spoken Language Understanding (SLU) systems are the building blocks of dialogue systems. The task of recognizing spoken language and interpreting the meaning behind it is the first step in creating these dialogue systems. A key task in SLU is intent recognition where given a spoken utterance we will recognize the intent behind that utterance (see fig \ref{fig:model}). 

In practice, this task is divided into Speech recognition and intent classification from the text output of the speech recognition system. This methodology is useful if we have enough data to train Automatic Speech Recognition systems and enough task data for the downstream classification task but it will not be helpful in low resourced settings.

In our context, by low resourced we mean two limitations; language scarcity which means we don't have enough data to train our speech to text systems, and task specific scarcity which means that for our task, data is not available to create a well behaved system.

End to End intent recognition systems can help solve this problem where we train a single model to carry out the task without sub-dividing the problem. \citet{anonymous2022on} show that the performance of using phoneme representation of the speech data and performing classification on them has better performance than using general voice representation techniques like wav2vec.

We build on this work by exploring the ability of both voice-space augmentation and phoneme-space augmentation to improve the performance of these phoneme-space models. We further exploit the embedding of phonemes located in phoneme recognition libraries like Allosaurus \citet{y4} to create data augmentation techniques that increase the amount of data available for training as well as make the models more robust. 

\section*{Main Contributions}

\begin{enumerate}
    \item Investigate the usage of pre-existing data augmentation techniques in low resource environments.
    \item Formulate novel phoneme specific data augmentation techniques to improve low-level classification performance.

\end{enumerate}

\section{Literature Review}

\subsection{Spoken Dialogue Systems}
The spoken dialogue system problem is concerned with developing systems that can naturally converse with humans. Spoken language systems typically consist of a Spoken Language Understanding (SLU) component, Dialogue State Tracking, Dialogue Policy Learning, and Natural Language Generation \citet{m1}. 

This formulation can be ignored in general task systems where we either use generative end-to-end models or retrieval-based models. \citet{m2} presented Meena, a generative, seq2seq based chat-bot that uses Evolved Transformers, a Neural Architecture Search model, as building blocks for their encoders and decoders. The chat-bot works by reducing the perplexity of the next token in the sentence, achieving a 23\% improvement over existing chat-bots in Sensibleness and Specificity Average (SSA) metric. \citet{m3} combine word sequence and utterance sequence information to generate their multi-view system conversational system, improving single word-view systems on context-specific conversational tasks. 

The Spoken Language Understanding task consists of Automatic Speech Recognition (ASR) and Natural Language Understanding (NLU) systems. As mentioned in \citet{m4}, ASRs are the bottleneck for developing SLUs for low resourced languages so this project will focus on this part of the system. 

\citet{m5} present an End-to-End SLU using a single trainable model, first training the model on predicting words and phonemes, then fine-tuning it to perform the full speech to intent task.

As mentioned in \citet{m6}, End-to-End ASRs work with a feature extractor to get features from the raw acoustic signal, an acoustic model that gives a probabilistic model of the features and a language model that matches word sequences with localized meanings. This formulation can be useful if applied to a space that has common symbol embeddings which can allow for ASR to work with minimal data.

\subsection{Intent Classification}
For the specific ASR application in this paper, the core task is intent classification. Intent Classification is a task concerned with understanding the intents of the user interacting with the system. For example, if the user utters the words: "I want the lights shut", a system such as Alexa \citet{ic2}, should be able to understand what the user's intent is, and in this case, act upon its knowledge of this intent. This task is often coupled with another task, Slot Filling. First, when performing intent classification, the correct semantic frame is identified, after that, at slot filling, these correct frames should be filled with the right value \citet{ic1}. These tasks could be performed jointly through a single model or separately with each task having its own parameters. 

For instance, in \citet{ic3}, researchers employ RNN based models in tackling the slot filling problem. Specifically, they use both Elman-type \citet{ic4} RNNs and Jordan-type \citet{ic5} RNNs. Their results show that bi-directional Jordan-type RNNs perform best on the task and outperform top baseline models at the time. 

In contrast, researchers \citet{ic6} use attention based RNN models to jointly tackle both tasks. They compare the same models when trained independently on each task versus when trained jointly on the tasks and report much improved performance of the latter. 

Apart from traditional RNN models and attention-based RNN models, these tasks were also tackled with transformers models. In \citet{ic6}, researchers show that BERT has the potential to tackle these tasks jointly and overcome the lack of proper training data often present in this context.

\subsection{Low Resource Speech Recognition}
Automatic Speech Recognition systems have various applications, including intent classification, in different spoken dialogue systems. In order to achieve a good performance, speech to text systems require a large amount of annotated speech data. However, the lack of annotated data is a major challenge to achieving these tasks in low-resourced languages.

In order to overcome the problem of scarcity of annotated data, \citet{y3} devised a new technique that uses acoustic phone units of the available data generated using Allosaurus library \citet{y4} for intent recognition systems. In the work \citet{y3} have used the system for a dummy banking app \citet{y2} that has 5 intents. Using the Naive Bayes classifier, they have achieved 0.83 accuracy with absolute discounting smoothing.

\citet{y1} used the same methodology where they generated phones from audio data and used the generated acoustic-phonetic units for intent classification. To perform classification, CNN+LSTM based architecture is proposed on two different language families. In addition, zero-shot performance for languages not present in the training set was shown. The work suggests that introducing a small amount of data for the language not present in the training set improves the performance significantly without affecting the performance of the other languages.

In another related work \citet{m4}, the authors used three different embeddings (Phone, Panphone, and Allo) extracted from Allosaurus \citet{y4} to perform intent classification on three different languages: English, Sinhala, and Tamil. This work simulates high, medium, and low resources on the languages. In their result, the authors have shown that their intent classification system improves SOTA intent classification accuracy by 2.11\% for Sinhala, and 7\% for Tamil and achieves competitive results in English. This work uses a 1-D dilated CNN based method to perform intent classification on the extracted embeddings.

\subsection{Intent Classification in Low Resource Settings}
As the task of intent classification is already challenging since it's hard to obtain sufficient and properly labeled training data, this issue is further amplified in the context of low-resource languages. 

In \citet{icl1}, researchers propose to tackle the problem of intent classification in Sinhala in a specific domain, and show a simple feed forward mlp with backpropagation can achieve acceptable accuracy. 

In \citet{icl2}, researchers propose utilizing an English language phoneme-based intent classifier to identify intents in utterances in Sinhala and Tamil. This is done by using a pre-trained English language based ASR model to identify such intents and show an accuracy of 80\% for as little as 30 minutes of Sinhala and Tamil data. 

Finally, in a recent paper \citet{icl3}, researchers tackle a very common issue in intent classification, often overlooked in the low resource context, that is, speaker variation. They propose the use of transfer learning to create speaker-invariant speech-to-intent classification system through the use of pre-trained acoustic models. The datasets are the Sinhala and Tamil datasets. They report that their method of using i-vector based augmentation improves performance when coupled with transfer learning.

\subsection{Data Augmentation for Speech Recognition}

In this section, different data augmentation methods for speech recognition are reviewed. We investigate the potential of a few in our low resource context, where it offers to mitigate the issue of having few labeled \_ or unlabelled \_ examples.

There are many varieties of data augmentation in the context of speech, and many algorithms that combine different methods together already exist. In \citet{g1}, researchers show that performing data augmentation with vocal tract length perturbation VTLP greatly improves results on speech recognition tasks. Researchers in \citet{g2} compare between speech perturbation, volume perturbation, and the combination of both in the context of a low resource language, namely, Turkish. They report that the combination yields the best results. 

Speech synthesis can also be combined with data augmentation to achieve better results. In \citet{g3}, researchers use an HMM-based speech synthesis system to create a dataset based on the BURNC corpus and compare different combinations of natural and synthesized speech. While their methods show improvement in English, they indicate it could yield similar results with LRLs. In \citet{aug0}, show that using speech synthesis in spoken language understanding systems and specifically to tackle intent classification shows success even when speech synthesis is made the sole source of training data.

In \citet{g4}, researchers from Google Brain present a new method that performs augmentation on the spectrogram itself, they call the method SpecAugment. The method takes the log mel spectrogram and first performs time warping, after that frequency masking is applied, and the last step is time masking. Another method presented in \citet{g5}, called MIxSpeech, where researchers train an Automatic Speech Recognition by taking a weighted combination of two different speech features, where recognition losses use the same combination weights. Their method is designed for low resource ASR and shows tremendous improvement on SpecAugment.

\section{Dataset}
In this paper, we use the Garbo dataset, which is a Belgian Dutch dataset \citet{renkens2014acquisition}. The dataset contains 36 commands (intents) spoken by 11 different speakers. The dataset is obtained from speakers commanding a service robot \citet{gupta2021intent}. This dataset is also divided into train, valid, and test splits.

\section{Model and Baseline Description \& Implementation}

Our baseline model is based on the work done by \citet{anonymous2022on}. The work presents a comparative analysis of a model built using speech features extracted from Word2Vec 2.0 and phonetic transcriptions extracted using Allosaurus \citet{y4}. After feature extraction, the embeddings are created using Convolutions Neural Networks from the phonetic features. Along with the embeddings, Long-Short Term Memory (LSTM) is used to learn the phonetic sequences in the embeddings. The result shows improvements from previous works and against word2vec 2.0 applied to the intent classification task for English and Flemish languages using the Grabo dataset for Flemish language \citet{renkens2014acquisition} and Fluent Speech Commands (FSC) dataset \citet{fsc} for the English language.  

From the two datasets used in the paper, different sizes of speakers, utterances per speaker, and intents were evaluated and studied using accuracy as a performance metric. Their experiments show that the new model outperformed previous works for the intent classification task. In addition, the work also shows that the model performs best when the number of utterances per speaker is kept to less than 4. For the number of utterances around 4 and 5, the performance saturates.

\begin{figure}[!ht]
    \centering
    \includegraphics[scale=0.75]{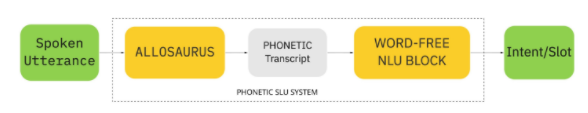}
    \caption{Baseline Model Architecture}
    \label{fig:base}
\end{figure}

Allosaurus is an open universal phoneme recognizer that can work on more than 200 languages. We use it to give us lower level representation of our utterances. The model then uses these representations as input to the intent recognition block which consists of a 1D CNN layer to act as an embedding layer and feature extractor. 

Another feature layer is the LSTM layer which takes the ReLU activated features from the previous layer and passes them on a uni-directional LSTM layer with no dropout. 

The final layer is a classification layer, a sigmoid-activated linear layer that outputs the prediction for the intent. Specifications for each layer are in table \ref{tab:arch}.

\begin{table}[h!]
\centering
\begin{tabular}{ c | c }
 \textbf{Model Parameter} & \textbf{Value} \\\hline 
 Embedding Size & 256 \\  
 CNN Kernel Size & 3 \\  
 No. CNN Filters & 256 \\  
 No. LSTM Layers & 1(or2) \\  
 LSTM Hidden Size & 256 \\  
 Batch Normalization & Flase  

\end{tabular}
\caption{Baseline Parameter Details}
\label{tab:arch}
\end{table}

This model presents State of the Art results in low-resourced intent classification on the Garbo dataset. It provides a 3\%, 32\% and 0.6\% increase over Wave2Vec models in this task in low (one speaker, one recording), medium (4 speakers, 4 recordings), and high (7 speakers, 7 recordings) data. We chose this model because of its performance and its simplicity over the other models.

\begin{figure}[!ht]
    \centering
    \includegraphics[scale=0.35]{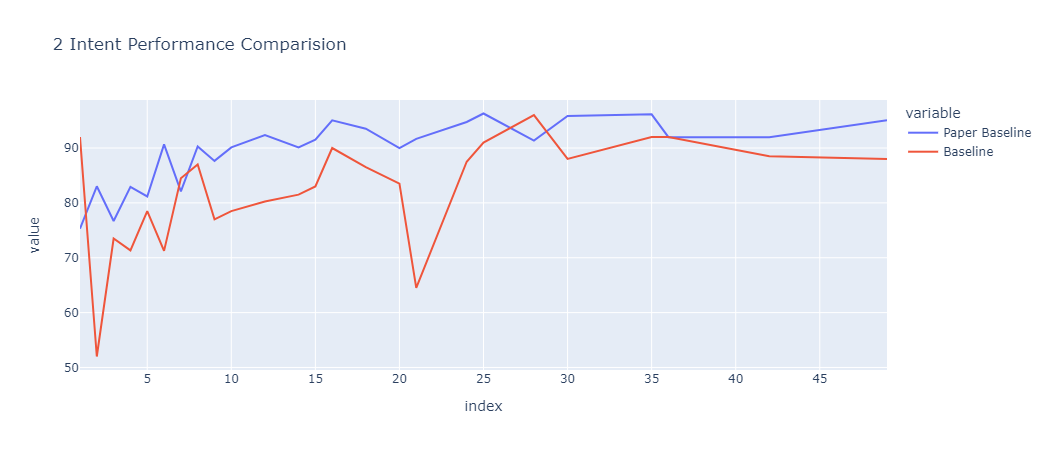}
    \caption{Baseline Comparison for Two Intents}
    \label{fig:base2base}
\end{figure}

\begin{figure}[!ht]
    \centering
    \includegraphics[scale=0.35]{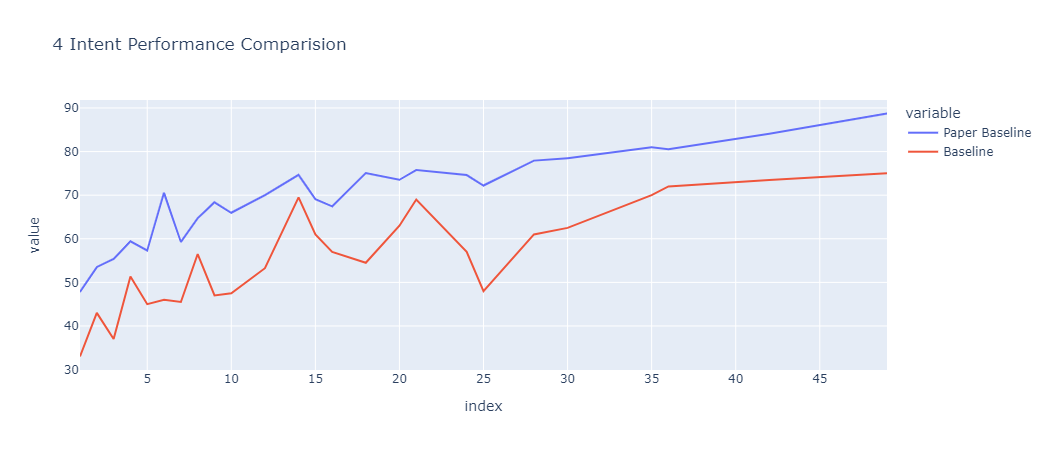}
    \caption{Baseline Comparison for Four Intents}
    \label{fig:base2base1}
\end{figure}

We have implemented the architecture presented in the baseline and started experimenting to reproduce the results that the \citet{y4} report in their paper. The table below shows the results. We decided to experiment with classifying 2 (I) intents on 1, 2, 6, and 7 (S) speakers with 1 to 7 (K) recordings per speaker. Below is a comparison between our implemented baseline and the original model from the paper. (see figures \ref{fig:base2base}, \ref{fig:base2base1}).

\section{Methodology for Augmentation}
\subsection{Voice Level}
Voice level augmentation refers to augmentation performed on the recordings before they're passed to the phoneme recognition. In this work, we have experimented with SpecAugment, speeding up the recording, and increasing the volume of the recording.

SpecAugment was implemented by converting the recording to a log mel-spectrogram tensor, and then using masking on both time and frequency dimensions, with a relatively small range. After that, the mel-spec is flattened and converted back to normal voice and then passed to the phoneme recognizer, that is because the phoneme recognizer used only accepts voice data. The resulting records were fairly audible and recognizable to the human ear, however, the phoneme recognizer ultimately failed to produce sensible transcripts even when many parameters were changed and experimented with while doing SpecAugment.

As for speeding up the recording, we experimented with factors between 1.2 to 2.5 and found that a factor of 1.6 is the most suitable for the phoneme recognizer to produce sensible transcripts. Finally, we experimented with increasing the volume within a range from 2 Decibels to 10 Decibels and found that an increase of 5 Decibels

In conclusion, we design a novel augmentation technique for the voice level using a randomized selection of speeding up the recording by a factor of 1.6 and then increasing the volume by 5 Decibels. 

\subsection{Phoneme Level}
Phoneme level augmentation refers to augmentation performed on the phoneme transcripts resulting from the phoneme recognizer. We design two novel techniques,  Allosaurus Noise Augmentation and Similar Phone Augmentation. 

\begin{figure}[!ht]
    \centering
    \includegraphics[scale=0.35]{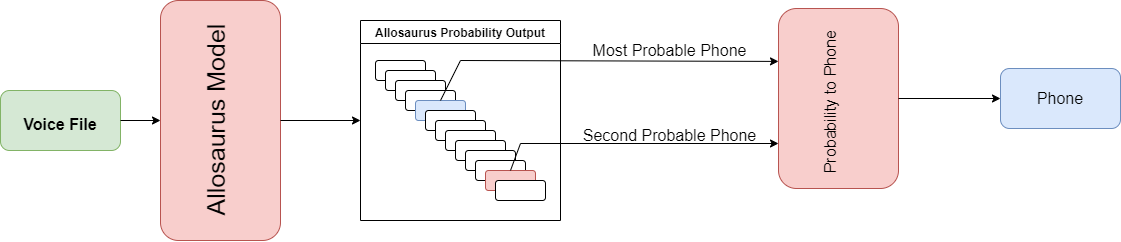}
    \caption{Allosaurus Noise Augmentation}
    \label{fig:allno}
\end{figure}

Allosaurus Noise Augmentation (Figure \ref{fig:allno}) employs the classification layer, the last layer in Allosaurus, which outputs a probability distribution over all the phones in the vocabulary. We insert noise to output from Allosaurus by using the second to last probable phoneme. We experiment with changing one or two phones in the phrase, clearly, changing the phone with a less probable phone can double the data available for training.

\begin{figure}[!ht]
    \centering
    \includegraphics[scale=0.35]{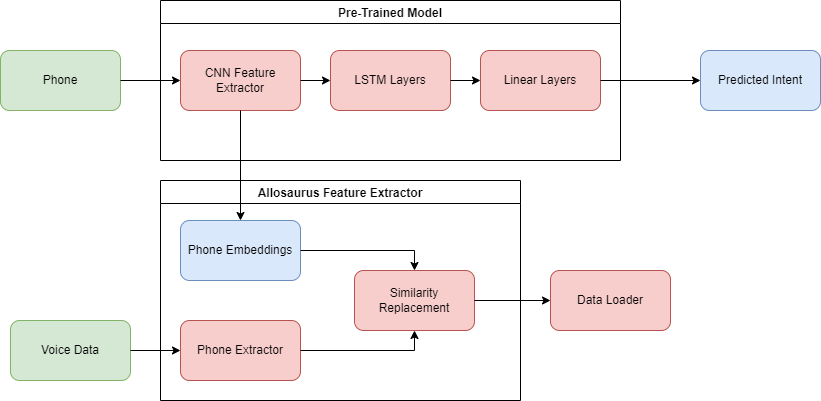}
    \caption{Similar Phone Augmentation}
    \label{fig:spa}
\end{figure}

Similar Phone Augmentation (Figure \ref{fig:spa}) uses cosine similarity on the feature representations extracted from the CNN output to map similar phones in the vocabulary to each other and further replace it while training. We believe this would make the model more robust against accents and subtle speech changes.  

\section{Results \& Discussion}
Results for the two intent classification are outlined in Figures \ref{fig:2p2i},\ref{fig:2pv2i}, \ref{fig:2s2i}, \ref{fig:2v2i}, \ref{fig:4a4i} shown below:
\begin{figure}[!ht]
    \centering
    \includegraphics[scale=0.35]{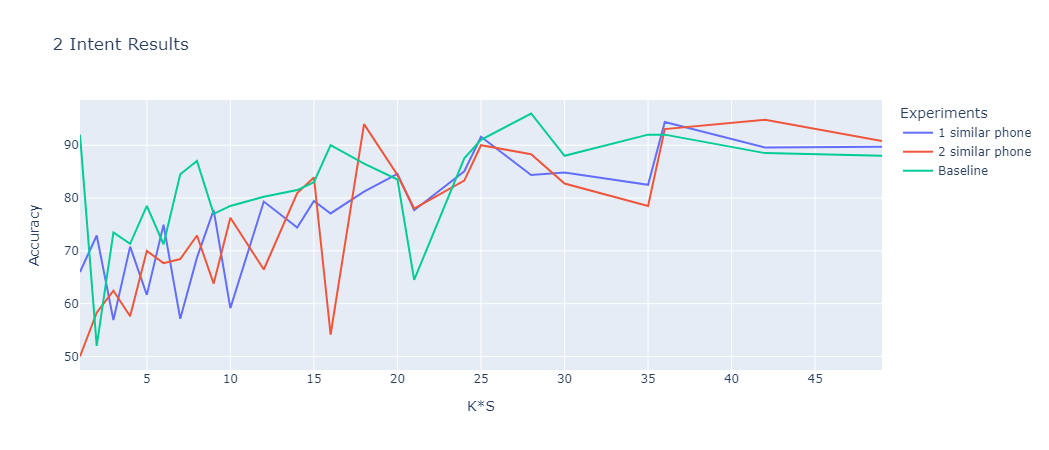}
    \caption{Similar Phone on 2 Intents}
    \label{fig:2s2i}
\end{figure}

\begin{figure}[!ht]
    \centering
    \includegraphics[scale=0.35]{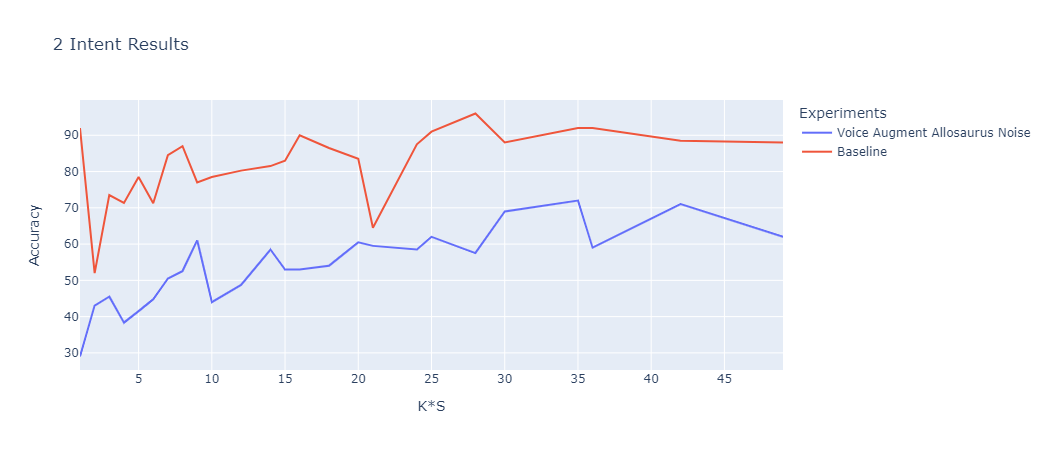}
    \caption{Voice Augment and Most Probable on 2 Intents}
    \label{fig:2pv2i}
\end{figure}

\begin{figure}[!ht]
    \centering
    \includegraphics[scale=0.35]{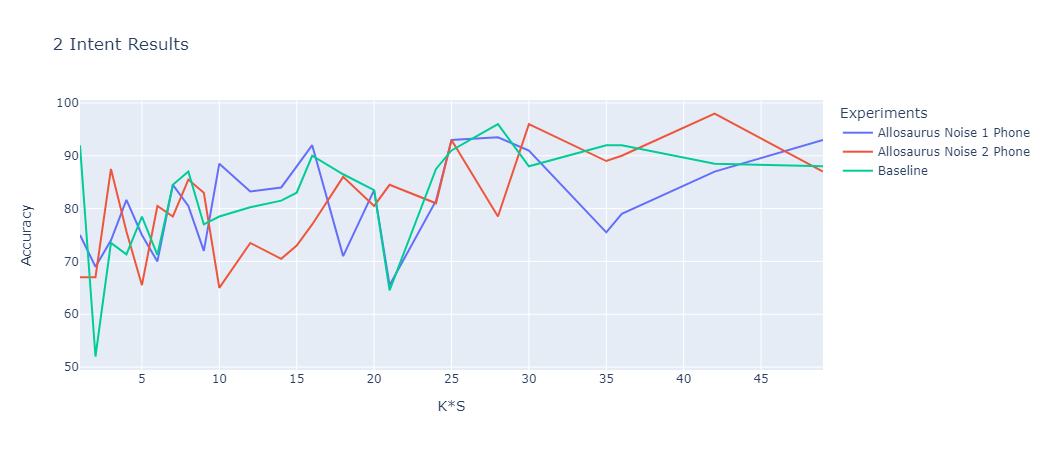}
    \caption{ Most Probable on 2 Intents}
    \label{fig:2p2i}
\end{figure}

\begin{figure}[!ht]
    \centering
    \includegraphics[scale=0.35]{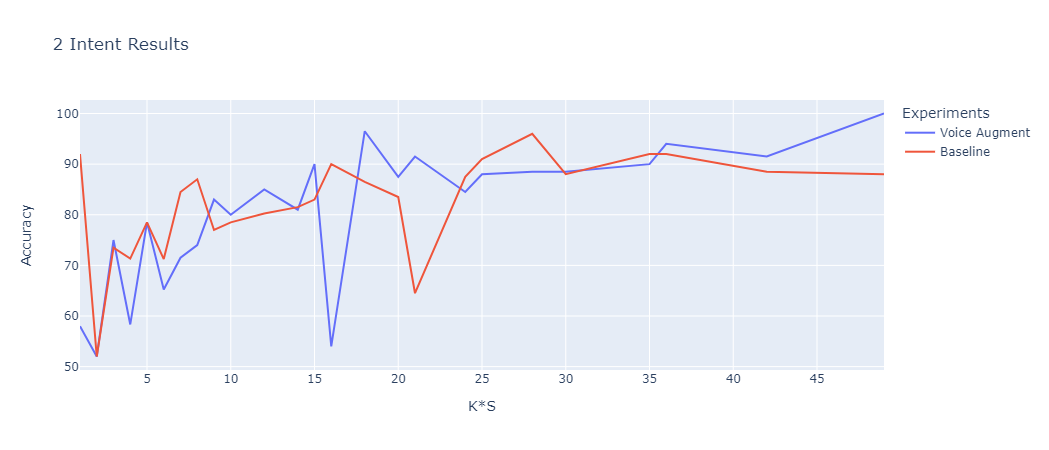}
    \caption{ Voice Augment on 2 Intents}
    \label{fig:2v2i}
\end{figure}

\begin{figure}[!ht]
    \centering
    \includegraphics[scale=0.35]{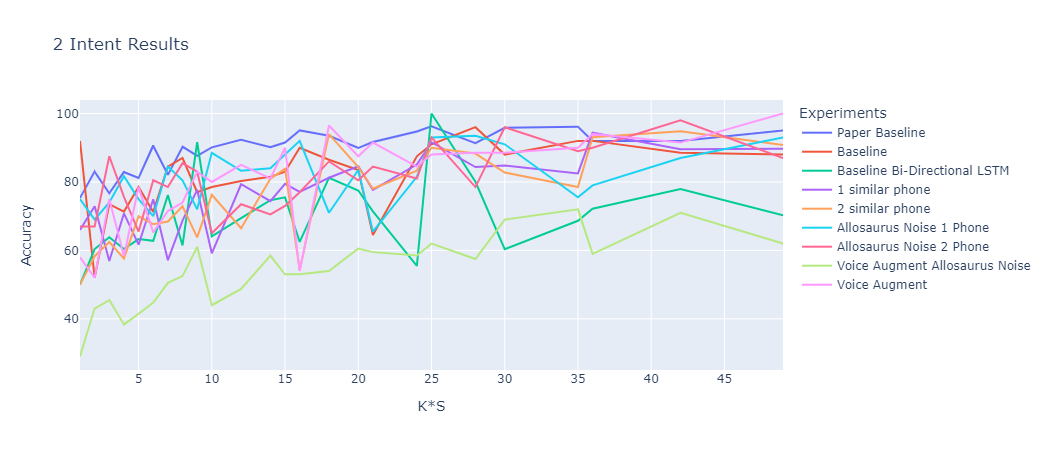}
    \caption{Comparison Between all Methods on 2 Intents}
    \label{fig:2a2i}
\end{figure}

\begin{figure}[!ht]
    \centering
    \includegraphics[scale=0.35]{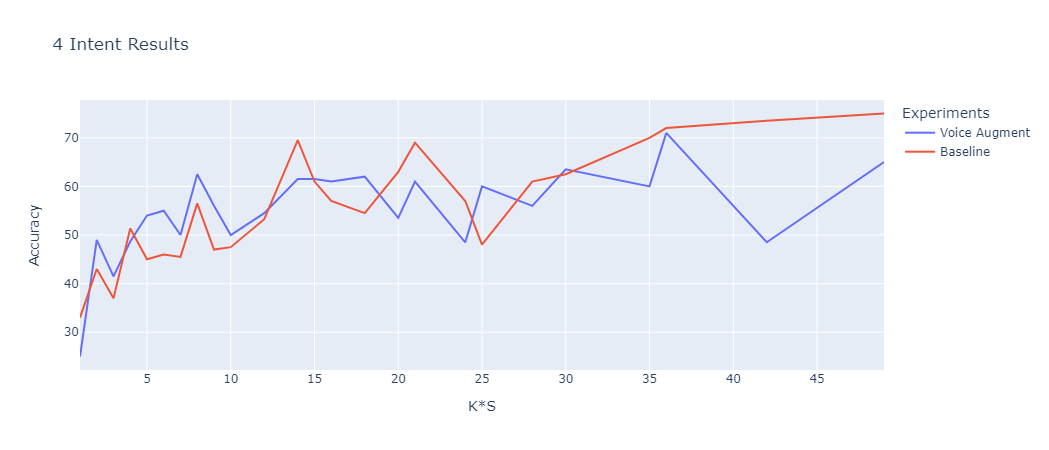}
    \caption{Voice Augment on 4 Intents}
    \label{fig:4v4i}
\end{figure}

As for the four intent classification tasks, results are outlined below in Figures \ref{fig:4a4i}, \ref{fig:4p4i}, \ref{fig:4pv4i}, \ref{fig:4v4i}. 
\begin{figure}[!ht]
    \centering
    \includegraphics[scale=0.35]{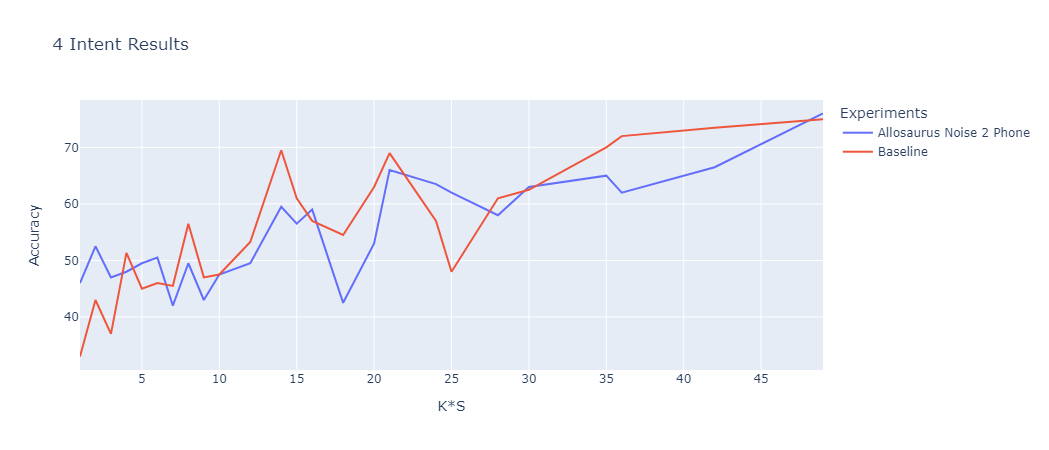}
    \caption{Most Probable on 4 Intents}
    \label{fig:4p4i}
\end{figure}

\begin{figure}[!ht]
    \centering
    \includegraphics[scale=0.35]{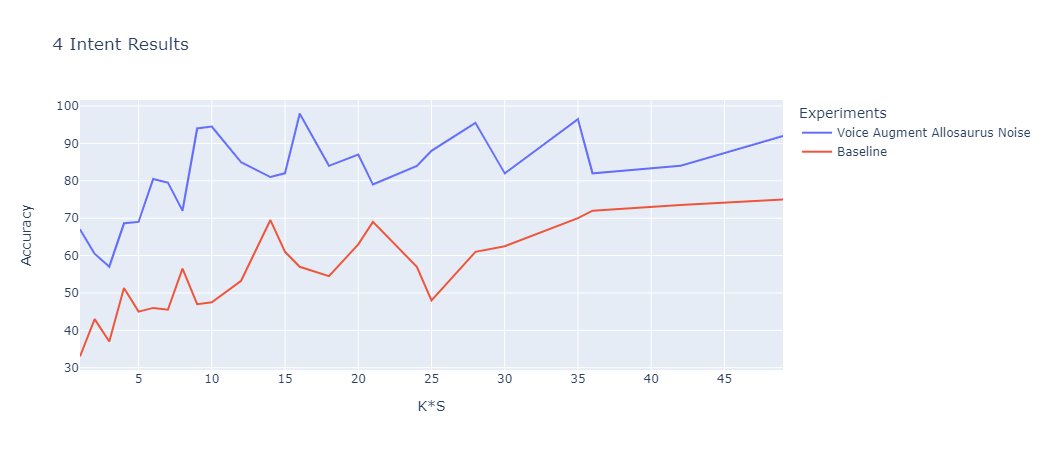}
    \caption{Voice Augment and Most Probable on 4 Intents}
    \label{fig:4pv4i}
\end{figure}

\begin{figure}[!ht]
    \centering
    \includegraphics[scale=0.35]{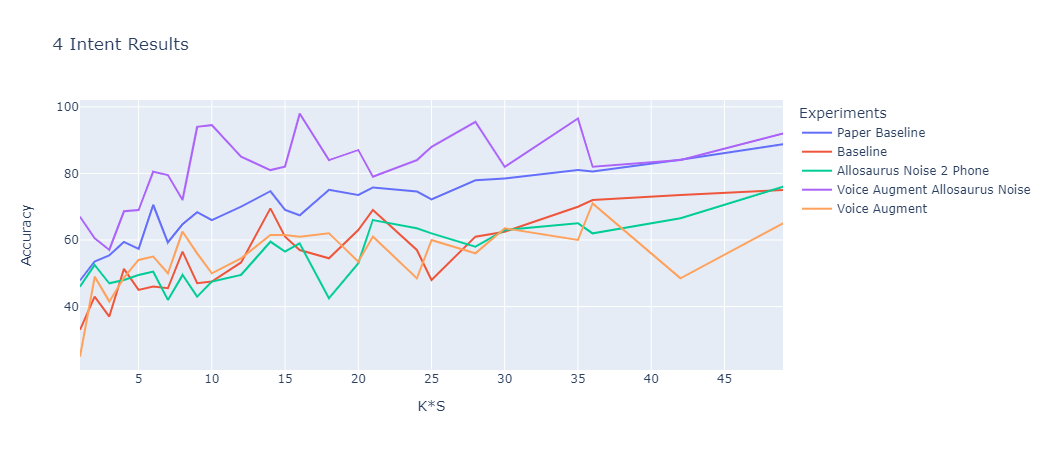}
    \caption{Comparison between all models on 4 Intents}
    \label{fig:4a4i}
\end{figure}

The results show that our novel augmentation methods outperform the baseline model, most notably, the combination of voice augment and noise on phonemes greatly outperforms the baseline model on the four intent classification task. 

As for why our model performs better on the four intent tasks, the basic intuition we have is that when working with four intents, where our augmentation performed well, the pool of data available was larger than for two intents, and that augmentation with the little amount of data the two intents task had simply introduced a lot of instability and noisiness during training.
\clearpage

\section{Conclusion and Future Work}
In this paper, we introduced two novel augmentation methods on two levels, the phoneme level, and the voice level. We find that our augmentation methods outperform the baseline model with a few choke points. 

In the future, we would like to explore the potential of using speech synthesis to diversify the training data, also, we aim to explore more feature extractors for the embeddings layer of the LSTM model.

\bibliographystyle{abbrvnat}
\bibliography{neurips_2019.bib}

\begin{thebibliography}{29}
\providecommand{\natexlab}[1]{#1}
\providecommand{\url}[1]{\texttt{#1}}
\expandafter\ifx\csname urlstyle\endcsname\relax
  \providecommand{\doi}[1]{doi: #1}\else
  \providecommand{\doi}{doi: \begingroup \urlstyle{rm}\Url}\fi

\bibitem[fsc()]{fsc}
The fluent speech commands (fsc) dataset.
\newblock URL
  \url{https://fluent.ai/fluent-speech-commands-a-dataset-for-spoken-language-understanding-research/}.

\bibitem[Adiwardana et~al.(2020)Adiwardana, Luong, So, Hall, Fiedel, Thoppilan,
  Yang, Kulshreshtha, Nemade, Lu, et~al.]{m2}
D.~Adiwardana, M.-T. Luong, D.~R. So, J.~Hall, N.~Fiedel, R.~Thoppilan,
  Z.~Yang, A.~Kulshreshtha, G.~Nemade, Y.~Lu, et~al.
\newblock Towards a human-like open-domain chatbot.
\newblock \emph{arXiv preprint arXiv:2001.09977}, 2020.

\bibitem[Buddhika et~al.(2018)Buddhika, Liyadipita, Nadeeshan, Witharana,
  Javasena, and Thayasivam]{icl1}
D.~Buddhika, R.~Liyadipita, S.~Nadeeshan, H.~Witharana, S.~Javasena, and
  U.~Thayasivam.
\newblock Domain specific intent classification of sinhala speech data.
\newblock In \emph{2018 International Conference on Asian Language Processing
  (IALP)}, pages 197--202. IEEE, 2018.

\bibitem[Elman(1990)]{ic4}
J.~L. Elman.
\newblock Finding structure in time.
\newblock \emph{Cognitive science}, 14\penalty0 (2):\penalty0 179--211, 1990.

\bibitem[Georgescu et~al.(2021)Georgescu, Pappalardo, Cucu, and Blott]{m6}
A.-L. Georgescu, A.~Pappalardo, H.~Cucu, and M.~Blott.
\newblock Performance vs. hardware requirements in state-of-the-art automatic
  speech recognition.
\newblock \emph{EURASIP Journal on Audio, Speech, and Music Processing},
  2021\penalty0 (1):\penalty0 1--30, 2021.

\bibitem[Gokay and Yalcin(2019)]{g2}
R.~Gokay and H.~Yalcin.
\newblock Improving low resource turkish speech recognition with data
  augmentation and tts.
\newblock In \emph{2019 16th International Multi-Conference on Systems, Signals
  \& Devices (SSD)}, pages 357--360. IEEE, 2019.

\bibitem[Gonfalonieri(2018)]{ic2}
A.~Gonfalonieri.
\newblock How amazon alexa works? your guide to natural language processing
  (ai), 2018.

\bibitem[Gupta(2022)]{anonymous2022on}
A.~Gupta.
\newblock On building spoken language understanding systems for low resourced
  languages.
\newblock In \emph{Submitted to 4th Workshop on NLP for Conversational AI},
  2022.
\newblock URL \url{https://openreview.net/forum?id=rrgMHxl-e-5}.
\newblock under review.

\bibitem[Gupta et~al.(2020{\natexlab{a}})Gupta, Rallabandi, and Black]{y1}
A.~Gupta, S.~K. Rallabandi, and A.~W. Black.
\newblock Mere account mein kitna balance hai? - on building voice enabled
  banking services for multilingual communities.
\newblock 2020{\natexlab{a}}.

\bibitem[Gupta et~al.(2020{\natexlab{b}})Gupta, Rallabandi, and Black]{y2}
A.~Gupta, S.~K. Rallabandi, and A.~W. Black.
\newblock Dummy banking app, 2020{\natexlab{b}}.
\newblock URL \url{https://awb.pc.cs.cmu.edu/}.

\bibitem[Gupta et~al.(2021{\natexlab{a}})Gupta, Deng, Kushwaha, Mittal, Zeng,
  Rallabandi, and Black]{gupta2021intent}
A.~Gupta, O.~Deng, A.~Kushwaha, S.~Mittal, W.~Zeng, S.~K. Rallabandi, and A.~W.
  Black.
\newblock Intent recognition and unsupervised slot identification for low
  resourced spoken dialog systems.
\newblock \emph{arXiv preprint arXiv:2104.01287}, 2021{\natexlab{a}}.

\bibitem[Gupta et~al.(2021{\natexlab{b}})Gupta, Li, Rallabandi, and Black]{y4}
A.~Gupta, X.~Li, S.~K. Rallabandi, and A.~W. Black.
\newblock Acoustics based intent recognition using discovered phonetic units
  for low resource languages.
\newblock 2021{\natexlab{b}}.

\bibitem[Ignatius and Thayasivam(2021)]{icl3}
A.~Ignatius and U.~Thayasivam.
\newblock Speaker-invariant speech-to-intent classification for low-resource
  languages.
\newblock In \emph{International Conference on Speech and Computer}, pages
  279--290. Springer, 2021.

\bibitem[Jaitly and Hinton(2013)]{g1}
N.~Jaitly and G.~E. Hinton.
\newblock Vocal tract length perturbation (vtlp) improves speech recognition.
\newblock In \emph{Proc. ICML Workshop on Deep Learning for Audio, Speech and
  Language}, volume 117, page~21, 2013.

\bibitem[Jordan(1997)]{ic5}
M.~I. Jordan.
\newblock Serial order: A parallel distributed processing approach.
\newblock In \emph{Advances in psychology}, volume 121, pages 471--495.
  Elsevier, 1997.

\bibitem[Karunanayake et~al.(2019)Karunanayake, Thayasivam, and
  Ranathunga]{icl2}
Y.~Karunanayake, U.~Thayasivam, and S.~Ranathunga.
\newblock Sinhala and tamil speech intent identification from english phoneme
  based asr.
\newblock In \emph{2019 International Conference on Asian Language Processing
  (IALP)}, pages 234--239. IEEE, 2019.

\bibitem[Li et~al.(2020)Li, Dalmia, Li, Lee, Littell, Yao, Anastasopoulos,
  Mortensen, Neubig, Black, and Metze]{y3}
X.~Li, S.~Dalmia, J.~Li, M.~Lee, P.~Littell, J.~Yao, A.~Anastasopoulos, D.~R.
  Mortensen, G.~Neubig, A.~W. Black, and F.~Metze.
\newblock Universal phone recognition with a multilingual allophone system.
\newblock 2020.

\bibitem[Liu and Lane(2016)]{ic6}
B.~Liu and I.~Lane.
\newblock Attention-based recurrent neural network models for joint intent
  detection and slot filling.
\newblock \emph{arXiv preprint arXiv:1609.01454}, 2016.

\bibitem[Louvan and Magnini(2020)]{ic1}
S.~Louvan and B.~Magnini.
\newblock Recent neural methods on slot filling and intent classification for
  task-oriented dialogue systems: A survey.
\newblock \emph{arXiv preprint arXiv:2011.00564}, 2020.

\bibitem[Lugosch et~al.(2019)Lugosch, Ravanelli, Ignoto, Tomar, and Bengio]{m5}
L.~Lugosch, M.~Ravanelli, P.~Ignoto, V.~S. Tomar, and Y.~Bengio.
\newblock Speech model pre-training for end-to-end spoken language
  understanding.
\newblock \emph{arXiv preprint arXiv:1904.03670}, 2019.

\bibitem[Lugosch et~al.(2020)Lugosch, Meyer, Nowrouzezahrai, and
  Ravanelli]{aug0}
L.~Lugosch, B.~H. Meyer, D.~Nowrouzezahrai, and M.~Ravanelli.
\newblock Using speech synthesis to train end-to-end spoken language
  understanding models.
\newblock In \emph{ICASSP 2020-2020 IEEE International Conference on Acoustics,
  Speech and Signal Processing (ICASSP)}, pages 8499--8503. IEEE, 2020.

\bibitem[Meng et~al.(2021)Meng, Xu, Tan, Wang, Qin, and Xu]{g5}
L.~Meng, J.~Xu, X.~Tan, J.~Wang, T.~Qin, and B.~Xu.
\newblock Mixspeech: Data augmentation for low-resource automatic speech
  recognition.
\newblock In \emph{ICASSP 2021-2021 IEEE International Conference on Acoustics,
  Speech and Signal Processing (ICASSP)}, pages 7008--7012. IEEE, 2021.

\bibitem[Mesnil et~al.(2013)Mesnil, He, Deng, and Bengio]{ic3}
G.~Mesnil, X.~He, L.~Deng, and Y.~Bengio.
\newblock Investigation of recurrent-neural-network architectures and learning
  methods for spoken language understanding.
\newblock In \emph{Interspeech}, pages 3771--3775, 2013.

\bibitem[Park et~al.(2019)Park, Chan, Zhang, Chiu, Zoph, Cubuk, and Le]{g4}
D.~S. Park, W.~Chan, Y.~Zhang, C.-C. Chiu, B.~Zoph, E.~D. Cubuk, and Q.~V. Le.
\newblock Specaugment: A simple data augmentation method for automatic speech
  recognition.
\newblock \emph{arXiv preprint arXiv:1904.08779}, 2019.

\bibitem[Patlan et~al.(2021)Patlan, Tripathi, and Korde]{m1}
A.~S. Patlan, S.~Tripathi, and S.~Korde.
\newblock A review of dialogue systems: From trained monkeys to stochastic
  parrots.
\newblock \emph{arXiv preprint arXiv:2111.01414}, 2021.

\bibitem[Renkens et~al.(2014)Renkens, Janssens, Ons, Gemmeke,
  et~al.]{renkens2014acquisition}
V.~Renkens, S.~Janssens, B.~Ons, J.~F. Gemmeke, et~al.
\newblock Acquisition of ordinal words using weakly supervised nmf.
\newblock In \emph{2014 IEEE Spoken Language Technology Workshop (SLT)}, pages
  30--35. IEEE, 2014.

\bibitem[Rygaard(2015)]{g3}
L.~V. Rygaard.
\newblock Using synthesized speech to improve speech recognition for
  lowresource languages.
\newblock \emph{Grace Hopper Celebration}, 2015.

\bibitem[Yadav et~al.(2021)Yadav, Gupta, Rallabandi, Black, and Shah]{m4}
H.~Yadav, A.~Gupta, S.~K. Rallabandi, A.~W. Black, and R.~R. Shah.
\newblock Intent classification using pre-trained embeddings for low resource
  languages.
\newblock \emph{arXiv preprint arXiv:2110.09264}, 2021.

\bibitem[Yan et~al.(2016)Yan, Song, and Wu]{m3}
R.~Yan, Y.~Song, and H.~Wu.
\newblock Learning to respond with deep neural networks for retrieval-based
  human-computer conversation system.
\newblock In \emph{Proceedings of the 39th International ACM SIGIR conference
  on Research and Development in Information Retrieval}, pages 55--64, 2016.

\end{thebibliography}

\end{document}